%% file: eacl2021.tex
\documentclass[11pt,a4paper,table]{article}
\usepackage[utf8]{inputenc}
\usepackage[hyperref]{eacl2021}
\usepackage{xcolor}
\usepackage[skip=0pt]{caption}
\usepackage{times}
\usepackage{dirtytalk}
\usepackage{latexsym}
\usepackage{amsmath}
\usepackage{mathtools}
\usepackage{tikz}
\usepackage{graphicx}
\usepackage{pgfplots}
\usepackage[T1]{fontenc}
\usepackage{inconsolata}
\usetikzlibrary{intersections}
\usepackage{array}
\usepackage{subcaption}
\usetikzlibrary{calc}
\usepackage{colortbl}
\usepackage{ctable}
\usepackage{float}
\usepackage{makecell}
\usepackage{tabularx}

\usepackage{ulem} 
\usetikzlibrary{arrows,shapes.geometric,positioning}
\usepgfplotslibrary{fillbetween}

\hypersetup{colorlinks=true,citecolor=blue}
\usepackage{microtype}

\aclfinalcopy 
\def\aclpaperid{1340} 

\title{\textsc{Enconter}: Entity Constrained Progressive Sequence Generation via Insertion-based Transformer}

\author{Lee-Hsun Hsieh \\\\\\
  \texttt{jesus255221@gmail.com}
  \\\And
  Yang-Yin Lee \\
  Singapore Management University \\ 
  Singapore \\
  \texttt{yylee@smu.edu.sg} \\\And
  Ee-Peng Lim \\\\\\
  \texttt{eplim@smu.edu.sg} \\
  }

\date{}

\begin{document}
\maketitle
\begin{abstract}

Pretrained using large amount of data, autoregressive language models are able to generate high quality sequences. However, these models do not perform well under hard lexical constraints as they lack fine control of content generation process.
Progressive insertion-based transformers can overcome the above limitation and efficiently generate a sequence in parallel given some input tokens as constraint.  These transformers however may fail to support hard lexical constraints as their generation process is more likely to terminate prematurely.  The paper analyses such early termination problems and proposes the \textsc{En}tity-\textsc{con}strained insertion \textsc{t}ransform\textsc{er} (\textsc{Enconter}), a new insertion transformer that addresses the above pitfall without compromising much generation efficiency. We introduce a new training strategy that considers predefined hard lexical constraints (e.g., entities to be included in the generated sequence).  Our experiments show that \textsc{Enconter} outperforms other baseline models in several performance metrics rendering it more suitable in practical applications.~\footnote{Our code is available at \url{https://github.com/LARC-CMU-SMU/Enconter}}
\end{abstract}

\section{Introduction}
\label{sec:introduction}

The field of Natural Language Generation (NLG)~\cite{gatt2018survey} has seen significant improvements in recent years across many applications such as neural machine translation~\cite{bahdanau2015neural}, text summarization~\cite{chopra2016abstractive}, poem generation~\cite{zugarini2019neural} and recipe generation~\cite{h2020recipegpt}.  Constrained text generation (CTG) is one of the challenging problems in NLG that is important to many real world applications but has not been well addressed.  CTG imposes input constraints which may be in the form of objects expected to exist in the generated text or rules over objects in the generated text~\cite{hokamp2017lexically}.  The objects here can be entities, phrases, predefined nouns, verbs, or sentence fragments. The constraints can be categorized into two types: (1)  \textit{Hard-constraints} which require mandatory inclusion of certain objects and complete compliance of given rules~\cite{post2018fast,hu2019improved, miao2019cgmh,welleck2019non,zhang2020pointer}; and (2) \textit{Soft-constraints} which allow the some constraint objects or rules to be not strictly enforced in the generated text~\cite{qin2019conversing,tang2019target}.  As autoregressive models generate tokens from left to right, they cannot easily support constraints involving multiple input objects, hard-constrained text generation therefore often requires non-autoregressive models.

Recently, \citet{zhang2020pointer} proposed a non-autoregressive hard-constrained text generation model (\textsc{Pointer}) that generates a text sequence in a progressive manner using an insertion-transformer~\cite{stern2019insertion}.  To train an insertion transformer to generate a missing token between every two tokens in an input sequence, the training data is prepared by masking ``less important'' tokens in the original text sequence in an alternating manner.  
The process is then repeated using the masked input sequence as the new original sequence, and further masking alternate tokens in it.  The process ends when the masked sequence meets some length criteria.  

While \textsc{Pointer} shows promising results, it does not consider hard constraints which involve entities that must be included in the generated sequence. Such entity constraint requirements are unfortunately prevalent in many applications.  For example, we may want to generate a job description with some given skills, or a food recipe with some given ingredients. 

A naive approach to the problem is to apply constraints on the \textsc{Pointer}'s masking strategy forcing it to keep entity tokens. We call this modified model \textsc{Pointer-E}. Although this allow entity information entering \textsc{Pointer-E}, another problem rises. \textsc{Pointer-E} suffers from \textit{cold start problem} which refers to the inability to generate meaningful tokens at the early stages of inference forcing the generation to end prematurely.  This issue can be attributed to the \textsc{Pointer-E}'s top-down masking strategy for training the insertion transformer and the tokens of input entities not evenly spread out across the sequence.  

To solve the cold start generation problem, we propose \textsc{Enconter} that incorporates bottom-up masking strategy. \textsc{Enconter} supports hard entity constraints, and encourages more meaningful tokens to be generated in the early stages of generation thus reducing cold start.  On top of that, we further introduce the balanced binary tree scheme~\cite{stern2019insertion} to reduce the number of stages in generation and to improve the efficiency of generation.

\begin{figure}[t!]
    \centering
    \begin{subfigure}[t]{0.48\textwidth}
    \centering
    \includegraphics[width=\textwidth]{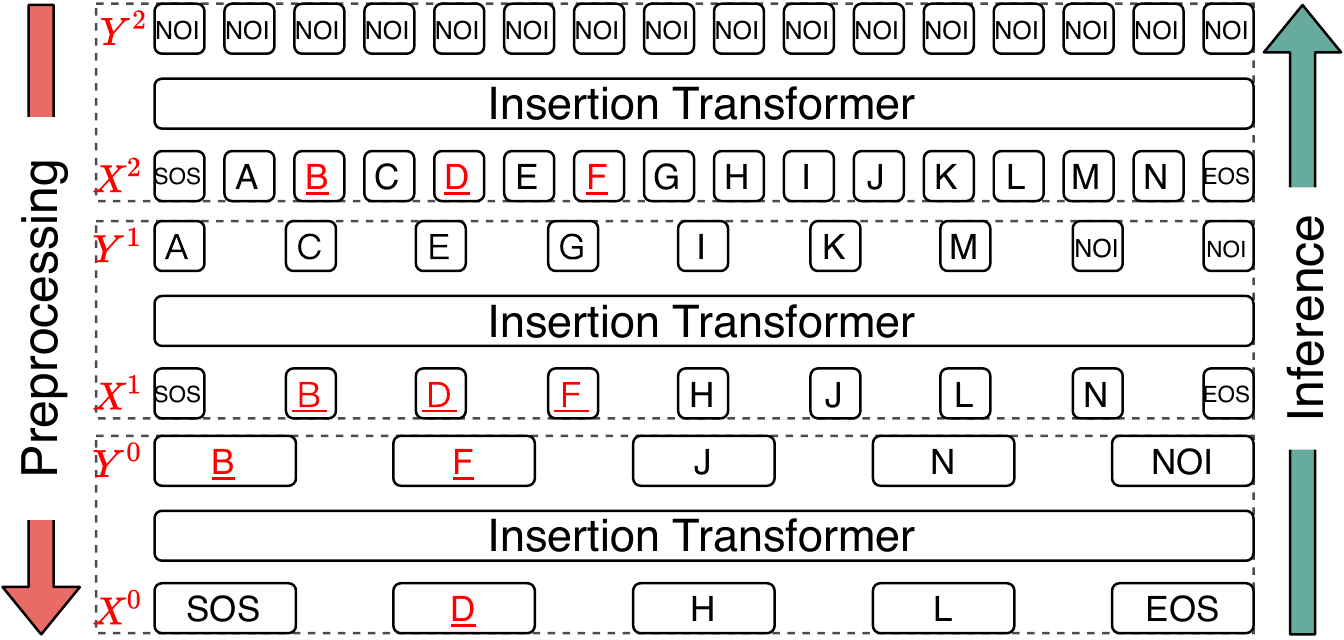}
    \caption{\textsc{Pointer} masking.}
    \label{fig:pointer_masking}
    \end{subfigure}
    \vspace*{4mm}
    
    \begin{subfigure}[t]{0.48\textwidth}
    \centering
    \includegraphics[width=\textwidth]{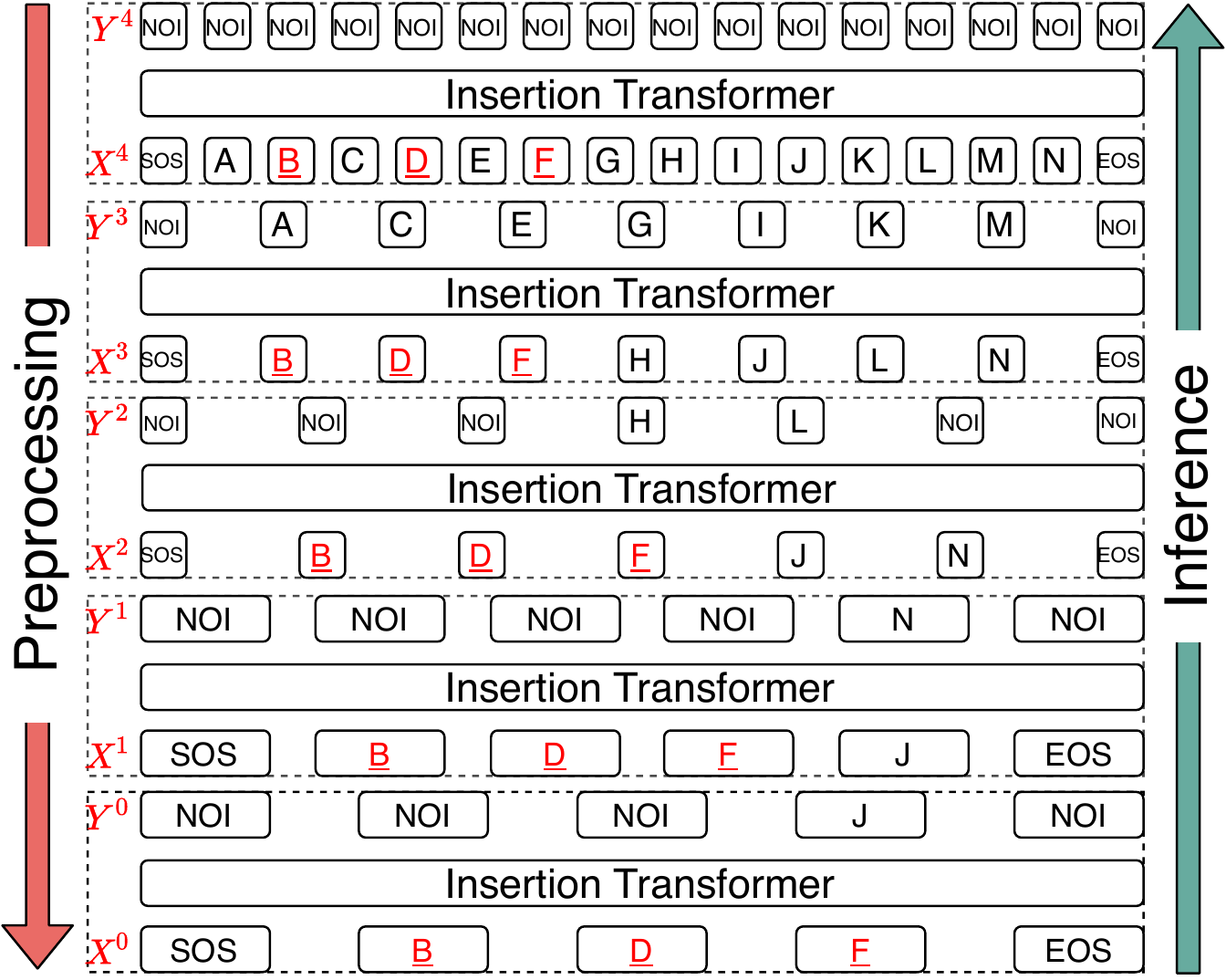}
    \caption{\textsc{Pointer-E} masking.}
    \label{fig:pointer_e_masking}
    \end{subfigure}
    \vspace*{4mm}
    
    \begin{subfigure}[t]{0.48\textwidth}
    \centering
    \includegraphics[width=\textwidth]{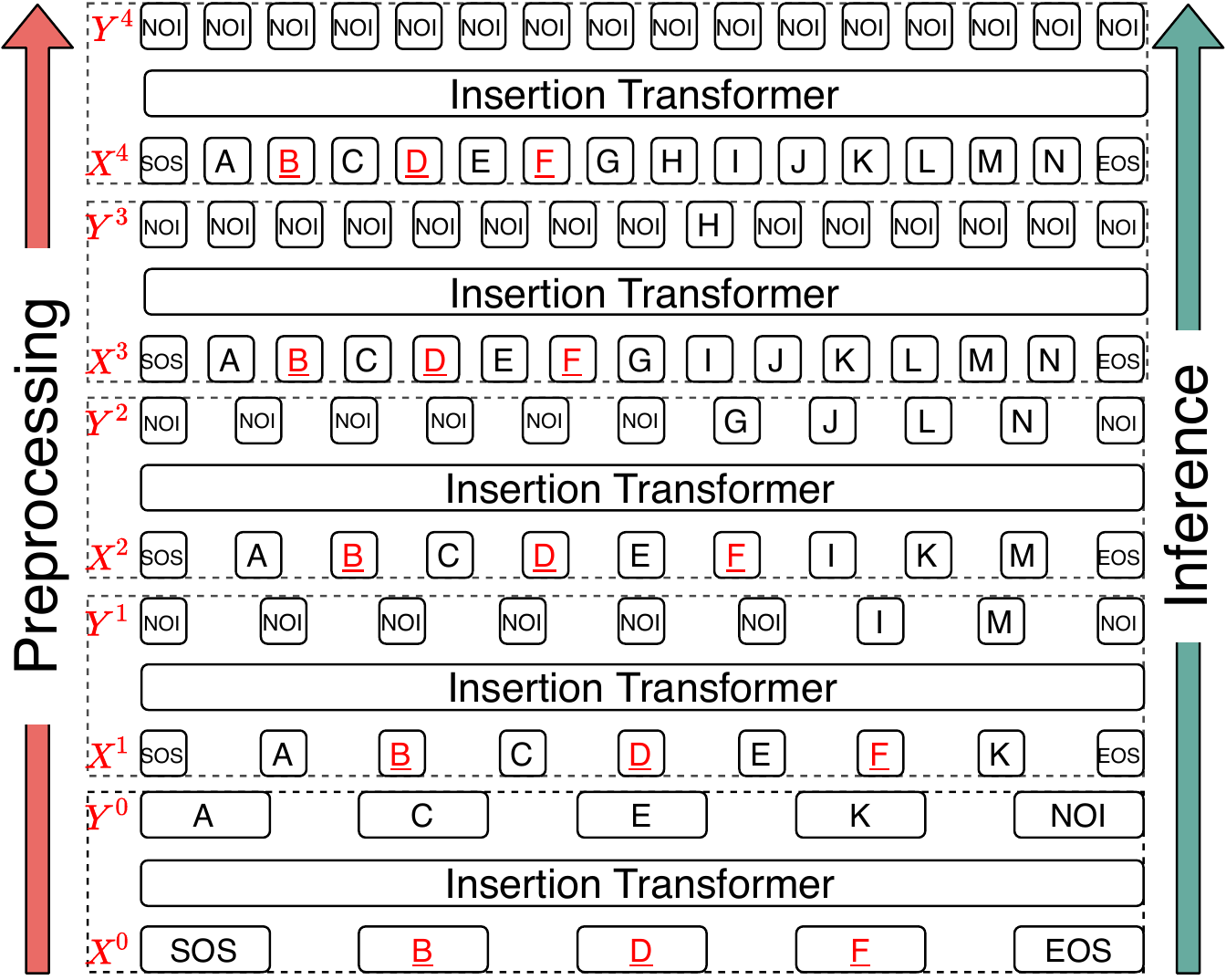}
    \caption{\textsc{Enconter} insertion.}
    \label{fig:enconter_masking}
    \end{subfigure}
    
    \caption{\textsc{Pointer}, \textsc{Pointer-E}, and \textsc{Enconter}  with original sequence $X=\{A,B,C,D,E,F,G,H,I,J,K,L,M,N\}$ where $B$,$D$, and $F$ are the tokens forming the entity constraints. The stopping criteria for \textsc{POINTER} is set to $n=3$.}
    \label{fig:maksing_figure}
\vspace{-3mm}
\end{figure}

\section{Entity Constrained Sequence Generation}
\label{sec:methodology}

In this section, we first describe the state-of-the-art \textsc{Pointer} model, its preprocessing of training data and inference process.  We highlight the pitfalls of the entity constrained variant of \textsc{Pointer}, \textsc{Pointer-E}.  We then present our proposed entity constrained insertion transformer called \textsc{Enconter}.

\subsection{\textsc{Pointer}}

\textsc{Pointer} adopts a progressive masking approach to train an insertion transformer. Let $X=\left\{x_1, x_2, \ldots, x_T\right\}$ denote  a a sequence where $x_t \in V$, where $T$ is the sequence length and $V$ is a finite vocabulary set.
Suppose $X$ is a training sequence, \textsc{Pointer} preprocesses it to obtain the training pairs $S=\left\{(X^k,Y^k)\middle| k\in\{K,\ldots,0\}\right\}$ using a progressive masking strategy. As shown in Figure~\ref{fig:pointer_masking}, in each stage $X^k$ represents the input sequence for stage $k$, and $Y^k$ represents the sequence of masked tokens to be inferred. $X^K$ is identical to the final training sequence $X^K=X$, and there should not be any additional tokens to infer. $X^0$ on the other hand represents the initial lexical constraints. In stage $k$, $Y^k$ are the tokens to be predicted between adjacent tokens of $X^k$.  A special no-insertion token $[NOI]$ is added to the vocabulary $V$ and used in $Y^k$ to indicate that no token is to be generated between adjacent tokens.  $Y^{K}$ is thus a sequence of all $[NOI]$'s indicating the end of generation. 
WordPiece~\cite{wu2016google} tokenization is applied in \textsc{Pointer}, and tokens split from the same word share the same score.

\noindent
\textbf{Token importance scoring} \textsc{Pointer} assigns each token $x_t \in X$ an importance score $\alpha_t$:
\begin{equation}
\label{eqn:tis}
    \alpha_t = \alpha_t^{TF-IDF} + \alpha_t^{POS} + \alpha_t^{YAKE},
\end{equation}
where $\alpha_t^{TF-IDF}$, $\alpha_t^{POS}$, and $\alpha_t^{YAKE}$ denote term frequency-inverse document frequency (TF-IDF), POS tag scores and YAKE~\cite{campos2020yake} keyword scores, respectively. These scores are normalized to [0,1]. $\alpha_t^{POS}$ is defined such that the scores of nouns and verbs are higher than those of other POS tags. 
The token importance scores are used to derive the masking pattern $Y^{k-1}$ of stage $k-1$ from $X^k$. 

\textsc{Pointer} adopts four criteria to derive $Y^{k-1}$ from $X^k$: (1) $Y^{k-1}$ can only include non-adjacent tokens in $X^k$;
(2) the number of tokens to be masked are maximized in each stage to make the model more efficient; 
(3) less important tokens are masked before more important ones and (4) A stopping criteria $n$ is defined. The algorithm stops when $|X^k| = n$.  Kadane's algorithm~\cite{gries1982note} has been use in \textsc{POINTER} to fulfill the criteria. Specifically, the algorithm selects as many unimportant tokens as possible to be masked while not masking two adjacent tokens. $X^0$ is automatically determined when $|X^k| = n$, it does not necessarily match the way the initial input sequence is provided by real world applications or users, including the entity constraints.

\noindent
\textbf{Inference} Given $X^0$ as input sequence, \textsc{Pointer} starts to infer $\hat{Y}^0$ and combines the two sequences to get $\hat{X}^1= \left\{\hat{x}_1^0, \hat{y}_{1}^0, \hat{x}_2^0, \hat{y}_{2}^0, \ldots, \hat{x}_{|\hat{X}^{0}|}^0, \hat{y}_{|\hat{X}^{0}|}^0\right\}$. If $\hat{y}_{t}^0$ happens to be $[NOI]$, it will be deleted and leaving only non-$[NOI]$ tokens in $\hat{X}^{1}$.  The process repeats until all the generated tokens in $\hat{Y}^k$ are $[NOI]$s.

As shown in Figure~\ref{fig:pointer_masking}, entities may not be preserved during the preprocessing steps and the lexical constraint $X^0$ is not guaranteed to cover entity  constraint $X^e$ even entity tokens are assigned high importance scores.  The trained \textsc{Pointer} therefore may not be able generate a sequence successfully when given entity constraints during the inference. We therefore propose some changes to \textsc{Pointer} to make it entity-aware.

\subsection{Entity Aware \textsc{Pointer} (\textsc{Pointer-E})}
\label{subsec:entity_aware_masking}

The entity-aware \textsc{Pointer} model, \textsc{Pointer-E}, adopts a different preprocessing approach.  
Let $X^e \subset X$ be an ordered sequence of entity tokens (e.g., the person names in a news document).  As $X^e$ is likely to be used as the initial generation input (i.e., $X^0=X^e$), \textsc{Pointer-E}'s preprocessing does not mask these entity tokens over the different preprocessing stages.  This way, the model is trained to focus on generating tokens around the entities. Such tokens form the context around the entities and context relating one entity to others.  We achieve such goal by ignoring the importance scores applied on entity tokens.  That is, we only compute $\alpha_t$ for $x_t \notin X^e$.

We then apply the \textsc{Pointer}'s masking strategy on the sub-sequence between every two entity tokens in $X$.  Suppose $(x_{i},  x_{j}) \subset X$ is a subsequence spanned by two entity tokens $\{x_l^e=x_{i}, x^e_{l+1}=x_{j}\} \in X^e$ where $l \in\{0,\ldots,|X^e|-1\}$.  Masking is applied on this subsequence iteratively until only $\{x_l^e, x_{l+1}^e\}$ are left:
\begin{equation}
    \begin{split}
    S= & \{(X^{K}=X, Y^{K}), \ldots, \\
       & (X^0=X^e, Y^0)\}.
    \end{split}
\end{equation}
As shown in Figure~\ref{fig:pointer_e_masking}, \textsc{Pointer-E} always picks the optimal masking patterns while preserving the entities.

\noindent
\textbf{Cold Start Problem} While \textsc{Pointer-E} is aware of entities, entities in $X^e$ may appear very close or very far from one another in the full sequence $X$, i.e., the gap between entities in the sequence $X$ can vary a lot.  Consider two sub-sequences $(x_i=x_l^e, x_j=x_{l+1}^e), (x_u=x_w^e, x_v=x_{w+1}^e) \subset X$ where $w, l \in (0, T_e - 1)$ and $w \neq l$.  Suppose $j - i \gg v - u$. The tokens between $(x_u, x_v)$ will then be masked out long before tokens in $(x_i, x_j)$ during preprocessing and training.  This results in \textsc{Pointer-E} trained to generate a lot of $[NOI]$s in $Y^k$ for small $k$'s.  Figure~\ref{fig:pointer_e_masking} depicts this cold start problem as entity tokens $B$, $D$ and $E$ are near one another in $X$.  As tokens between them are masked in early stages, the masked sequences in stages 0 and 1, $Y^0$ and $Y^1$, contain many $[NOI]$ tokens. \textsc{Pointer-E} trained with such data will therefore lack the ability to generate meaningful tokens in-between these entity tokens. In the worst case, \textsc{Pointer-E} simply generates all $[NOI]$ tokens and ends the generation prematurely which is known as the \textit{cold start problem}.   

To better show the problem, we define:
\begin{equation}
    NOI\;ratio = \frac{\#[NOI]\;\text{in $Y^k$}}{\#tokens\;\text{in $Y^k$}}.
\end{equation}
A clear problem of high $NOI\;ratio$ is that $Y^k$ is very similar to $Y^{k+1}$. When $NOI\;ratio=1$, the generation will end,  In cases where $NOI\;ratio$ is very high for masked sequences in early stages, say $Y^0$,  the trained \textsc{Pointer-E} will more likely infer from $X^0$ all $[NOI]$'s for $\hat{Y}^0$ and end the generation process.  To address this, we need to re-examine the top-down masking stratey used in \textsc{Pointer} and \textsc{Pointer-E}.  

\subsection{\textsc{Enconter}}
\label{sec:enconter}
\noindent
In this section, we propose \textsc{Enconter} which adopts a bottom-up masking strategy to overcome the cold start problem.  There are two variants: \textsc{Greedy Enconter} and \textsc{BBT-Enconter}.

\noindent
\textbf{\textsc{Greedy Enconter}} Different from \textsc{Pointer-E}, we now construct training pairs $S$ from $X$ by setting $X^0$ to be $X^e$:
\begin{equation}
    \begin{split}
    S=&\{(X^{0}=X^e, Y^0), (X^{1}, Y^{1}), \ldots,\\
    & (X^K=X, Y^{K})\},
    \end{split}
\end{equation}
where $Y^k$ represents the sequence of masked tokens to be inserted into $X^k$ to form $X^{k+1}$.  Similar to \textsc{Pointer}, $Y^K$ contains $[NOI]$'s only.  For every two adjacent tokens $\{x^k_t, x^k_{t+1}\} \in X^k$ where $t \in \left\{0,\ldots, |X^k| - 1\right\}$, we insert a mask token.  Let $\{x^k_t = x_i, x^k_{t+1} = x_j\}$ and $(x_i, x_j)$ be the span of $(x_i, x_j)$ in $X$.  If $i+1=j$, the mask token is $[NOI]$.  Otherwise, we select a token $x_{t'}$ from $(x_i, x_j)$ with maximum importance score $\alpha_{t'}$ within $(x_i, x_j)$ as the mask token.  The sequence $Y^k$ is formed after we go through all the $t$'s.  By inserting $Y^k$ into $X^k$, we obtain the next sequence $X^{k+1}$.  The iterative process stops when all the tokens to be inserted are $[NOI]$s.  This method \textit{\textsc{Greedy Enconter}} greedily selects the token with maximum importance score in the span to be generated in a bottom up insertion (or unmasking) process.  By forcing more non-$[NOI]$ tokens to be included in $Y^0$ and $Y^k$ of small $k$'s, \textit{Greedy Enconter} achieves lower $NOI\;ratio$ in the early stages of inference.  Experimentally, we find that the cold start problem is eliminated.  

\noindent
\textbf{Balanced binary tree \textsc{Enconter} (\textsc{BBT-Enconter})} To further improve the efficiency of \textsc{Greedy Enconter}, we incorporate balanced binary tree~\cite{stern2019insertion} into \textsc{Enconter} to bias the masking of tokens to be those near the center of the unobserved subsequence of tokens.  BBT reward is added to the importance score function as follows. Suppose $x_i$ and $x_j$ are two adjacent tokens in $X^k$, and $(x_i, x_j)$ represents the corresponding subsequence in $X$.  We define the distance $d_p$ for token $x_p \in (x_i, x_j)$ as:
\begin{equation}
    d_p = min(p - i,j - p).
\end{equation}
We use a softmax function to compute the reward for weighted score based on $d_p$:
\begin{equation}
    w_p = \frac{exp(d_p/\tau)}{\sum^j_{k=i}exp(d_k/\tau)}.
\end{equation}
The weights in the span are then normalized to $[0,1]$.  Then the importance score is defined as:
\begin{equation}
    \alpha_p = 
        w_p \cdot  (\alpha_p^{TF-IDF} + \alpha_p^{POS} + \alpha_p^{YAKE}).
    \label{eq.bbt}
\end{equation}
The construction of $S$ is almost the same as \textsc{Greedy Enconter}. The only difference is the new importance score function defined by Eq.~\ref{eq.bbt}.  This proposed model, known as \textit{\textsc{BBT-Enconter}}, will predict the center and semantically important token in $X$ between two adjacent tokens of $X^k$.

\begin{figure*}[htb]
    \centering
    \includegraphics[width=\textwidth]{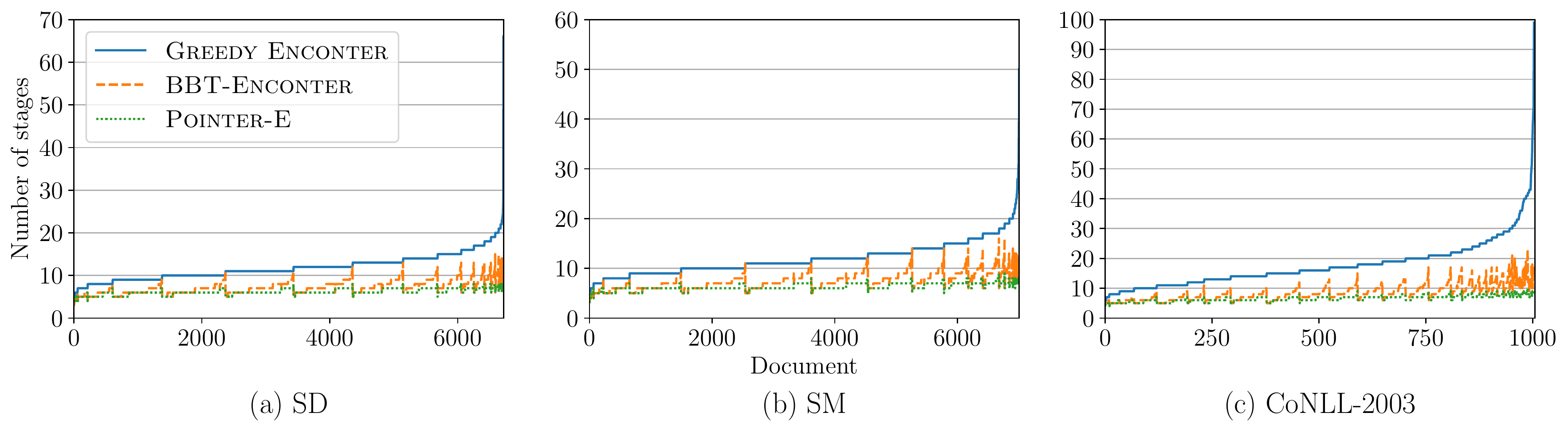}
\vspace{-3mm}
\caption{Number of stages of each document in SD, SM, and CoNLL-2003 (sorted).}
\label{fig:compare_stage}
\vspace{-3mm}
\end{figure*}

\subsection{Models with Entity Span Aware Inference Option (ESAI)}

So far, all above-mentioned models assume that each entity consists of one single token.  In real world use cases, an entity may contain more than one token.  Without any control during the inference process, it is possible for other tokens to be generated in-between tokens of the same entity.  For example in Table~\ref{tab:SM_gen_exm}, "\uline{Group Consolidation}" may be split into "handling \uline{Group} s project / \uline{Consolidation}".  To avoid inserting any tokens in between any multi-token entity, we introduce the \textit{entity span aware inference} option to the inference process of \textsc{Pointer-E} and \textsc{Enconter} to force the inference of $\hat{Y}^k$ to always generate $[NOI]$ in between the tokens of the multi-token entities.  After applying ESAI, the multi-token entities will remain unbroken duing the generation process.

\section{Empirical Analysis of \textsc{Pointer-E} and \textsc{Enconter}}

In this section, we conduct an analysis of the data preprocessing step in \textsc{Pointer-E}, \textsc{Greedy Enconter} and \textsc{BBT-Enconter}. Our objective is to empirically evaluate the characteristics of training data generated for the two models.  We have left out \textsc{Pointer} as it is inherently not entity-aware and \textsc{Pointer-E} is its entity-aware variant. We first present the two datasets used in this study.

\begin{figure*}[htb]
    \raggedright
    \begin{subfigure}[b]{0.29\textwidth}
    \centering
    \input{pics/SDnoi.tikz}
    \vspace*{-5mm}
    \caption{SD}
    \end{subfigure}
    \hspace{1em}
    \begin{subfigure}[b]{0.29\textwidth}
    \centering
    \input{pics/SMnoi.tikz}
    \vspace*{-5mm}
    \caption{SM}
    \end{subfigure}
    \hspace{1em}
    \begin{subfigure}[b]{0.29\textwidth}
    \centering
    \input{pics/CoNLLnoi.tikz}
    \vspace*{-5mm}
    \caption{CoNLL-2003}
    \end{subfigure}
    
\caption{Mean and standard deviation of the ratio of inserted/masked [NOI] tokens in each stage.  All x axis are capped to 15 stages.  The original maximum number of stages of (a), (b), and (c) are 67, 51, and 100, respectively.}
\label{fig:compare_inserted_noi}
\vspace{-3mm}
\end{figure*}
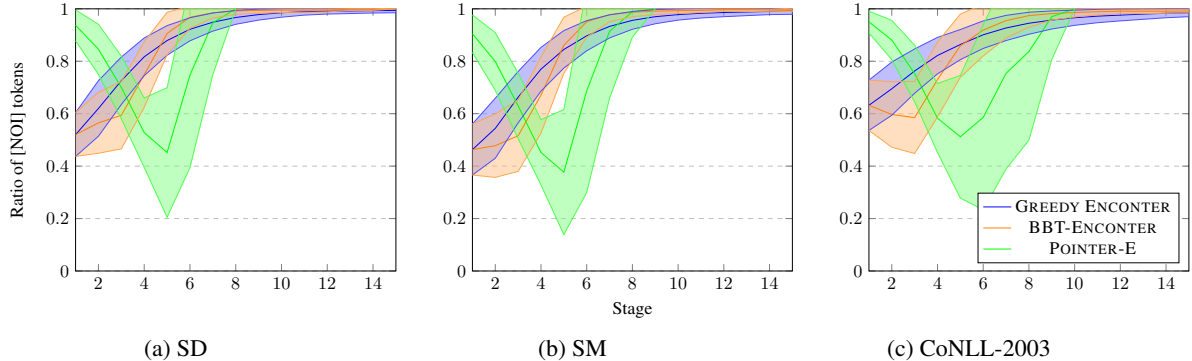

\subsection{Datasets}
\label{subsec:dataset}
\textbf{CoNLL-2003}~\cite{tjong2003introduction}: We select the English version which contains 1,393 news articles labeled with four named entity types: persons, locations, organizations and names of miscellaneous. Training and development sets are used to train the model. Documents having more than 512 tokens by wordpiece tokenizer used in BERT~\cite{devlin-etal-2019-bert} are discarded to ensure that the whole document can fit into the models.

\noindent
\textbf{Jobs}: This is a job post dataset collected from Singapore's Jobsbank~\footnote{https://www.mycareersfuture.sg/}. The dataset consists of 7,474 job posts under the software developer occupation (SD) and 7,768 job posts under the sales and marketing manager occupation (SM). We extract the requirement section of these job posts as the text sequences to be generated.  For each requirement text sequence (or document), we use a dictionary of skills to annotate the skill and job related entities in the sequence.

The detailed information of the datasets can be found in Table~\ref{tbl:dataset_summary}.  Table~\ref{tbl:dataset_summary} reveals that \textsc{Pointer-E} has much higher $NOI\;ratio$ than \textsc{Enconter} in all the datasets.

\begin{table}[h!]
\centering
\small
\begin{tabular}{l|r|r|r}
\hline & \textbf{CoNLL} & \textbf{SM} & \textbf{SD} \\ \hline
\#training docs & 1,004 & 6,715 & 7,006 \\
\#testing docs & 231 & 754 & 761 \\
Avg length & $220.7$ & $99.4$ & $121.1$ \\
Avg entities & $24.6$ & $24.4$ & $27.7$ \\
\hline
\#training pairs &&& \\
\textsc{Pointer-E} & 6,557 & 43,913 & 41,343 \\
\textsc{Greedy Enconter} & 17,694 & 83,587 & 79,467 \\
\textsc{BBT-Enconter} & 8,492 & 52,609 & 48,625 \\\hline
$NOI\;ratio$ of $Y^0$&&&\\
\textsc{Pointer-E}  & 0.820 & 0.904 & 0.936 \\
\textsc{Greedy Enconter} & 0.546 & 0.463 & 0.519 \\
\textsc{BBT-Enconter} & 0.546 & 0.463 & 0.519 \\
\hline
\end{tabular}
\caption{\label{tbl:dataset_summary} Summary of the datasets. \#training pairs refers to the total number of training pairs derive from each dataset}
\vspace{-3mm}
\end{table}

\label{sec:analysis_of_td_and_bu}

\subsection{Analysis of NOI ratio and Stage Counts}
We first analyse the ratio of $[NOI]$ tokens inserted or masked in every stage of the \textit{training data}.  Figure~\ref{fig:compare_inserted_noi} shows the mean together with one standard deviation of the \textsc{Pointer-E}, \textsc{Greedy Enconter} and \textsc{BBT-Enconter} for each dataset.  X-axis is in log scale.  Note we add $1$ to the stage number for showing log scale (e.g., the $10^0$ in the figure indicates the ratio of $[NOI]$ tokens in $Y^0$).  From Figure~\ref{fig:compare_inserted_noi}, we find all datasets share a few similar characteristics, namely: (1) for \textsc{Pointer-E}, the $[NOI]$ ratio is quite high in the first few stages, and drops when the stage is higher.  A sudden increase of the ratio to $1$ is due to the ending sequence consists all $[NOI]$'s; (2) for \textsc{Enconter} the $[NOI]$ ratio is low in the first few stages, and slowly increase to $1$.  The result shows \textsc{Enconter} can learn to generate balance proportion of $[NOI]$ and non-$[NOI]$ tokens in the first few stages, and also learn not to generate to many non-$[NOI]$ tokens when approaching the end of the generation process.

Figure~\ref{fig:compare_stage} shows the number of stages each training document requires under different models.  The numbers are sorted according to the following priority: \textsc{Greedy Enconter}, \textsc{Pointer-E}, then \textsc{BBT-Enconter}.  Since \textsc{BBT-Enconter} incorporates the binary tree reward scheme, it is able to perform insertion in the {\color{black} middle stages} more efficiently compare to \textsc{Greedy Enconter}.  This helps to lower the total number of stages required to derive training pairs.

\begin{table*}[t!]
\centering
\footnotesize
\setlength\tabcolsep{3pt}
\begin{tabular}{|l|r||rr|rr|r||r|rr|r|r||r|r|}
\hline
Method & Recall & \multicolumn{2}{|c|}{NIST} & \multicolumn{2}{|c|}{BLEU} & MTR & Entropy & \multicolumn{2}{|c|}{DIST} &   PPL & AvgLen & failure & AvgSteps \\
{} & {} & N-2 & N-4 & B-2 & B-4 & {} & E-4 & D-1 & D-2 &   {} & {} & {} & {} \\\hline \hline
\multicolumn{14}{|l|}{Baselines}\\ \hline
GPT-2  &   0.70 &   1.38 &   1.39 &   0.13 &   0.08 &   0.19 &      4.91 &   0.16 &   0.57 & \bf{35.7} &  201.4 &    \bf{0.00} &   256.84 \\
\textsc{Pointer-E}                &   0.98 &   0.72 &   0.72 &   0.08 &   0.04 &   0.19 &      3.63 &   0.22 &   0.65 & 285.7 &   88.9 &    0.35 &     4.18 \\
\textsc{Pointer-E} (+ESAI)         &   \bf{1.00} &   0.63 &   0.64 &   0.08 &   0.04 &   0.18 &      3.54 &  \bf{0.23} &   \bf{0.67} & 337.1 &   81.4 &    0.34 &     3.84 \\\hline \hline
\multicolumn{14}{|l|}{\textsc{Enconter}}\\ \hline
Greedy                &   0.96 &   1.95 &   1.96 &   0.19 &   0.09 &   \bf{0.25} &      \bf{4.99} &   0.16 &   0.58 & 112.1 &  192.5 &   \bf{0.00} &    17.56 \\
Greedy (+ESAI)         &   \bf{1.00} &   \bf{1.99} &   \bf{2.00} &   \bf{0.20} &   \bf{0.10} &   \bf{0.25} &      4.95 &   0.16 &   0.59 & 111.7 &  181.9 &    \bf{0.00} &    16.48 \\
BBT      &   0.94 &   1.83 &   1.84 &   0.19 &   \bf{0.10} &   0.23 &      4.87 &   0.18 &   0.62 & 154.9 &  161.1 &    \bf{0.00} &     8.19 \\
BBT (+ESAI) &   \bf{1.00} &   1.87 &   1.87 &   \bf{0.20} &   \bf{0.10} &   0.24 &      4.84 &   0.18 &   0.62 & 150.0 &  156.7 &    \bf{0.00} &     8.26 \\
\hline \hline

Human        & - & - & - & - & - & - &      4.99 &   0.20 &   0.66 &  53.0 &  202.1 & - & - \\\hline
\end{tabular}
\caption{CoNLL-2003 result}
\label{tab:CoNLL_result}
\vspace{-3mm}
\end{table*}

\begin{table*}[t!]
\centering
\footnotesize
\setlength\tabcolsep{3pt}
\begin{tabular}{|l|r||rr|rr|r||r|rr|r|r||r|r|}
\hline
Method & Recall & \multicolumn{2}{|c|}{NIST} & \multicolumn{2}{|c|}{BLEU} & MTR & Entropy & \multicolumn{2}{|c|}{DIST} & PPL & AvgLen & failure & AvgSteps \\
{} & {} & N-2 & N-4 & B-2 & B-4 & {} & E-4 & D-1 & D-2 &   {} & {} & {} & {} \\\hline\hline
\multicolumn{14}{|l|}{Baselines}\\ \hline
GPT-2 &   0.63 &   1.42 &   1.42 &   0.13 &   0.09 &   0.20 &      4.63 &   0.05 &   0.30 & \bf{66.3} &  123.2 &    \bf{0.00} &   168.44 \\
\textsc{Pointer-E}                &   0.99 &   1.69 &   1.70 &   0.20 &   0.12 &   0.28 &      3.84 &   \bf{0.08} &   \bf{0.46} & 901.2 &   73.4 &    0.38 &     5.01 \\
\textsc{Pointer-E} (+ESAI)         &   \bf{1.00} &   1.71 &   1.72 &   0.21 &   0.12 &   0.28 &      3.85 &   \bf{0.08} &   \bf{0.46} & 966.4 &   74.1 &    0.34 &     4.91 \\\hline\hline
\multicolumn{14}{|l|}{\textsc{Enconter}}\\ \hline
Greedy                &   0.99 &   3.31 &   3.33 &   0.40 &   0.28 &   0.41 &      4.53 &   0.07 &   0.37 & 124.3 &  111.1 &    \bf{0.00} &     9.80 \\
Greedy (+ESAI)         &   \bf{1.00} &   3.31 &   3.33 &   0.40 &   0.28 &   0.40 &      4.52 &   0.07 &   0.37 & 125.2 &  109.5 &    \bf{0.00} &     9.79 \\
BBT        &   0.99 &   \bf{3.59} &   \bf{3.62} &   \bf{0.45} &   \bf{0.34} &   \bf{0.44} &      4.53 &   0.07 &   0.38 & 135.6 &  110.6 &    \bf{0.00} &     5.86 \\
BBT (+ESAI) &   \bf{1.00} &   3.55 &   3.57 &   0.44 &   \bf{0.34} &   \bf{0.44} &      \bf{4.54} &   0.07 &   0.38 & 137.7 &  111.5 &    \bf{0.00} &     5.92 \\
\hline
Human         &   - &  - &  - &  - &  - &  - &  4.66 &   0.08 &   0.41 &  90.7 &  125.1 & - & -  \\\hline
\end{tabular}
\caption{SD result}
\label{tab:SD_result}
\vspace{-3mm}
\end{table*}

\begin{table*}[t!]
\centering
\footnotesize
\setlength\tabcolsep{3pt}
\begin{tabular}{|l|r||rr|rr|r||r|rr|r|r||r|r|}
\hline
Method & Recall & \multicolumn{2}{|c|}{NIST} & \multicolumn{2}{|c|}{BLEU} & MTR & Entropy & \multicolumn{2}{|c|}{DIST} &   PPL & AvgLen & failure & AvgSteps \\
{} & {} & N-2 & N-4 & B-2 & B-4 & {} & E-4 & D-1 & D-2 &   {} & {} & {} & {} \\\hline\hline
\multicolumn{14}{|l|}{Baselines}\\ \hline
GPT-2 &   0.72 &   1.50 &   1.51 &   0.15 &   0.10 &   0.23 &      \bf{4.40} &   0.05 &   0.32 & \bf{101.0} &   96.4 &    \bf{0.00} &   127.48 \\
\textsc{Pointer-E}                &   0.98 &   1.32 &   1.32 &   0.17 &   0.10 &   0.26 &      3.46 &   \bf{0.09} &   \bf{0.48} & 2447.7 &   52.7 &    0.34 &     4.89 \\
\textsc{Pointer-E} (+ESAI)         &   \bf{1.00} &   1.26 &   1.26 &   0.16 &   0.09 &   0.25 &      3.42 &   \bf{0.09} &   \bf{0.48} & 2535.7 &   53.3 &    0.38 &     5.07 \\\hline\hline
\multicolumn{14}{|l|}{\textsc{Enconter}}\\ \hline
Greedy                &   0.99 &   2.48 &   2.49 &   0.31 &   0.20 &   0.36 &      4.21 &   0.07 &   0.40 & 153.9 &   82.2 &    \bf{0.00} &     9.75 \\
Greedy (+ESAI)         &   \bf{1.00} &   2.44 &   2.45 &   0.31 &   0.20 &   0.36 &      4.19 &   0.07 &   0.40 & 147.4 &   80.2 &    \bf{0.00} &     9.62 \\
BBT         &   0.98 &   \bf{2.73} &   \bf{2.74} &   \bf{0.34} &  \bf{0.24} &   \bf{0.38} &      4.26 &   0.07 &   0.41 & 161.1 &   83.8 &    \bf{0.00} &     6.04 \\
BBT (+ESAI) &   \bf{1.00} &   2.69 &   2.70 &   \bf{0.34} &   0.23 &   \bf{0.38} &      4.25 &   0.07 &   0.41 & 157.5 &   83.6 &    \bf{0.00} &     6.05 \\
\hline

Human         & - & - & - & - & - & - &      4.45 &   0.08 &   0.43 & 104.3 &  101.6 & - & - \\\hline

\end{tabular}
\caption{SM result}
\label{tab:SM_result}
\vspace{-6mm}
\end{table*}

\section{Experiment}
\label{sec:experiment}

\subsection{Models for Comparison}

\noindent
\textbf{GPT-2}~\cite{radford2019language} GPT-2 can be used to conduct conditional generation as well (soft-constraints).  For a training sequence $X$ together with its entities $X^e$, we concatenate $X^e$ with $X$ to form a training sequence $\{X^e, X\}$.  $X^e$ is then served as a control code sequence to guide GPT-2 in the generation of $X$.  We fine-tune the \textit{GPT-2 small} pretrained by huggingface~\footnote{https://huggingface.co/} with $10^{-5}$ learning rate.  Warmup and weight decay are applied.  10 epochs are used for fine-tuning.

\noindent
\textbf{\textsc{Pointer-E}, \textsc{Greedy Enconter}, and \textsc{BBT-Enconter}}: We use BERT~\cite{devlin-etal-2019-bert} as the underlying insertion transformer for all these models similar to that of \textsc{Pointer}.  Specifically, we use the \textit{bert-based-cased} pretrained by huggingface.  BERT with language model head is fine-tuned on all the training pairs to obtain the models.  Learning rate is set to $10^{-5}$ with warmup and weight decay. 10 epochs are used for fine-tuning. 
For \textsc{Pointer-E}, \textsc{Greedy Enconter}, and \textsc{BBT-Enconter}, top-k (top-$20$) sampling method is used to derive $\hat{Y}^k$.  For GPT-2, we feed in the $\hat{X}^e$ and let GPT-2 generate the following tokens until reaching the end-of-generation token.

\begin{table}[h!]
    \centering
    \small
    \begin{tabular}{>{\raggedright\arraybackslash}p{7.25cm}} \hline
    \textbf{\textsc{Greedy Enconter}:} \\
     Degree / \textcolor{blue}{\uline{ACCA}} / \textcolor{blue}{\uline{CIMA}} / CA / \textcolor{blue}{\uline{CFA}} / F \& B / \textcolor{blue}{\uline{Group Consolidation}} experience \\
     Degree in \textcolor{blue}{\uline{General Accounting}} \\
     Good track record in \textcolor{blue}{\uline{IFRS}}RSM, \textcolor{blue}{\uline{Fixed Assets}}, \textcolor{blue}{\uline{IFRS}} \\\hline
    \textbf{\textsc{Greedy Enconter} ESAI:} \\
     * * Degree / \textcolor{blue}{\uline{ACCA}} / \textcolor{blue}{\uline{CIMA}} / CA / \textcolor{blue}{\uline{CFA}} / CA / CAPA / Singapore \textcolor{blue}{\uline{Group Consolidation}} / Management / \textcolor{blue}{\uline{General Accounting}} experience \\
 * Experience in \textcolor{blue}{\uline{IFRS}} or preferred \\
 * Experience in \textcolor{blue}{\uline{Fixed Assets}} Management \\
 * Extensive experience in \textcolor{blue}{\uline{IFRS}} \\\hline
    \textbf{\textsc{BBT-Enconter}:} \\
    Degree or \textcolor{blue}{\uline{ACCA}} / \textcolor{blue}{\uline{CIMA}} or \textcolor{blue}{\uline{CFA}} qualifications \\
 YEARSPAN'experience in handling \textcolor{red}{\uline{Group}} s project / \textcolor{red}{\uline{Consolidation}} / Good Inteconor / \textcolor{blue}{\uline{General Accounting}} \\
 Knowledge of \textcolor{blue}{\uline{IFRS}}PAN, \textcolor{blue}{\uline{Fixed Assets}} ( \textcolor{blue}{\uline{IFRS}}, etc ) \\\hline
    \textbf{\textsc{BBT-Enconter} ESAI:} \\
     Minimum Degree / \textcolor{blue}{\uline{ACCA}} / \textcolor{blue}{\uline{CIMA}}, \textcolor{blue}{\uline{CFA}} or equivalent \\
 Minimum of YEARSPAN of experience in Marketing and \textcolor{blue}{\uline{Group Consolidation}} and \textcolor{blue}{\uline{General Accounting}} \\
 Knowledge of \textcolor{blue}{\uline{IFRS}} ) and \textcolor{blue}{\uline{Fixed Assets}} ( \textcolor{blue}{\uline{IFRS}} ) \\\hline
    \textbf{GPT-2:} (missing: \textcolor{red}{\uline{IFRS}}, \textcolor{red}{\uline{CFA}}, \textcolor{red}{\uline{Group Consolidation}}, \textcolor{red}{\uline{Fixed Assets}}) \\
    Job Requirements : \\
     - Degree in \textcolor{blue}{\uline{General Accounting}} / \textcolor{blue}{\uline{ACCA}} Qualification \\
     - At least YEARSPAN of applicable working experience in similar capacity \\
     - Must be able to multi-task and handle different priorities simultaneously \\
     - \textcolor{red}{\uline{General accounting}} knowledge will be advantageous \\
     Interested applicant, kindly send in your CPA or \textcolor{blue}{\uline{CIMA}} reference number to EMAIL \\
     EA Licence number : LICENSENUM \\
     Registration number : REGNUM \\\hline
    \textbf{Human:} \\
     Professional Qualifications : Bachelors Degree \\
     Qualified with a professional financial body ( ICAEW / ICPA / \textcolor{blue}{\uline{ACCA}} / \textcolor{blue}{\uline{CIMA}} / \textcolor{blue}{\uline{CFA}} etc ) \\
     Specialist Knowledge / Skills : \textcolor{blue}{\uline{Group Consolidation}} \textcolor{blue}{\uline{General Accounting}} \textcolor{blue}{\uline{IFRS}} \textcolor{blue}{\uline{Fixed Assets}} Industry Experience \\
     Experience : YEARSPAN post qualified with extensive \textcolor{blue}{\uline{IFRS}} experience and industry experience \\\hline
    \end{tabular}
    \caption{A generated example from SM dataset.  \textsc{Pointer-E} and \textsc{Pointer-E} ESAI are not shown since they failed to generate at first step.}
    \label{tab:SM_gen_exm}
\end{table}

\subsection{Evaluation Metrics}
We evaluate the models using a few criteria, namely: recall of entities, quality with respect to human crafted text, diversity, fluency, cold start, and generation efficiency.  We measure recall of entity constraints by the proportion of entity tokens found in the generated text.  Even without ESAI, the recall metric will allow to compare the recall ability of models.  Besides recall, we also consider BLEU~\cite{papineni2002bleu}, METEOR (MTR)~\cite{lavie2007meteor} and NIST~\cite{doddington2002automatic}, which are common metrics for evaluating the quality of generated text against human craft text.  We compute the BLEU-2 (B-2) and BLEU-4 (B-4) which are n-gram precision-based metrics.  For the BLUE based evaluation metric NIST, we compute the NIST-2 (N-2) and NIST-4 (N-4).  To measure the diversity of generation, Entropy~\cite{zhang2018generating} and Distinction~\cite{li2016diversity} are used. Entropy-4 (E-4) is defined as the frequency distribution of unique 4-gram terms. Dist-1 (D-1) and Dist-2 (D-2) are used to derive distinct  n-grams in the generated text. We also utilize pretrained language model to measure fluency. Perplexity (PPL) is calculated using pretrained GPT-2~\cite{radford2019language} without fine-tuning. The lower the perplexity is, the more fluent the generation is (based on GPT-2).  ``AvgLen'' is the averaged word counts of the generated sequence.  ``failure'' indicates the proportion of test sequences that fail to be generated at the first step (i.e., $\hat{Y}^0$ are all $[NOI]$'s).  Finally, ``AvgSteps'' shows the average number of steps for the model to complete the generation.  Note for GPT-2, the AvgSteps is based on tokens, while the AvgLen is based on words.

\subsection{Experiment Results}
\label{subsec:experiment_result}
Tables~\ref{tab:CoNLL_result}, \ref{tab:SD_result}, and \ref{tab:SM_result} show the results of different models on the different datasets. On recall, GPT-2, due to its inability to enforce hard lexical constraints, yields the worst recall.
For non-autoregressive models without ESAI, they still achieve high recall. Nevertheless, the high recall of \textsc{Pointer-E} is ``contributed by'' relatively high failure ratio ("failure") as recall is 1 even when the model fails to generate anything in the first stage. In other words, \textsc{Pointer-E} suffers from cold start problem.  \textsc{Greedy Enconter} and \textsc{BBT-Enconter}, in contrast, enjoy both good recall and zero failure ratio. With ESAI option, all non-autoregressive models can achieve perfect recall without much additional generation steps.  However, this option does not reduce the high failure ratio of \textsc{Pointer-E}.  On generation quality compared with human crafted text, \textsc{Greedy Enconter} and \textsc{BBT-Enconter} outperform all other models by NIST, BLEU, and MTR. This suggests that \textsc{Enconter} models learn the context of entities better compared to other models. On generation diversity,  \textsc{Pointer-E} again has the highest diversity largely due to its high failure ratio. Finally, we discuss the efficiency of models measured by AvgSteps. The autoregressive nature of GPT-2 makes it the least efficient model among all. \textsc{Pointer-E}'s ability to optimize masking patterns makes it the most efficient model. With balance binary tree reward, \textsc{BBT-Enconter} is able to finish its generation in fewer iterations than \textsc{Greedy Enconter}.

\noindent
\textbf{Case example}
Table~\ref{tab:SM_gen_exm} shows a case example from Jobs SM dataset. The entities of the given constraint are underlined. Invalid entities generated are colored in red, while the remaining ones are colored in blue.  There are three types of invalid cases. First, the case of entity is not the same as specified.  Second, the entity is not recalled in the generation.  Third, the entity has its tokens separated by some other token(s).  In this example, \textsc{Pointer-E} and \textsc{Pointer-E} ESAI terminate their generations prematurely. They fail to perform generation at the very first stage.

\section{Related Work} 
\label{sec:related-work}

Recent years have witnessed significant success using autoregressive~\cite{dai2015semi,peters2018deep,Radford2018ImprovingLU} generative models to conduct conditional generation on various tasks. CTRL~\cite{keskarCTRL2019} uses control codes trained together with large amount of data to control the content to be generated. RecipeGPT~\cite{h2020recipegpt} takes ingredients as a series of control and trains the generation of recipe text. PPLM~\cite{Dathathri2020Plug} directly steers the pretrained language by a bag-of-words model or simple linear discriminator. The above models in their own ways  gain certain level of control over the content generation process. However, they do not provide a mechanism to directly enforce some lexical constraints in the final generation. Non-monotonic sequence generation~\cite{welleck2019non} is designed to perform hard lexical constrains generation based on binary tree structure. By leveraging level-order and inorder traversal of binary tree, the model allows text to be generated non-monotonically. Although the results from non-monotonic generation models seem promising, they do not perform token generation in parallel and the tree structure governing the generation process may produce many unused tokens during the generation.

The emergence of non-autoregressive language model provides another approach to support hard lexical constraints.  Insertion transformer~\cite{stern2019insertion} uses transformer architecture with balanced binary tree loss to perform insertion-based generation. KERMIT~\cite{chan2019kermit} is proposed as a structure to unify insertion transformers. Levenshtein transformer~\cite{gu2019levenshtein} further introduces deletion as an action to take during generation. Our \textsc{Enconter} models differ from these previous models as they are not designed to support any lexical constraints, including entity constrains. 

\section{Conclusions}
\label{sec:conclusions-and-future-works}

Constrained text generation is an important task for many real world applications.  In this paper, we focus on hard entity constraints and the challenges associated with enforcing them in text generation. Our analysis of the state-of-the-art insertion transformers reveals issues, namely, cold start problems and inefficient generation. We therefore propose two insertion transformer models, \textsc{Greedy Enconter} and \textsc{BBT Enconter}, that use a bottom-up preprocessing strategy to prepare training data so as to eliminate the cold start problem caused by top-down preprocessing strategy.  \textsc{BBT Enconter} further incorporates a balanced tree reward scheme to make the generation process more efficient.  Through experiments on real world datasets, we show that the two models outperform the strong baselines, \textsc{Pointer-E} and GPT2, in recall, quality and failure rate while not compromising much generation efficiency. For future research, it will be interesting to consider more diverse constraints (e.g., soft constraint, rules, etc.) and user interaction in the generation process to expand the scope of applications that can benefit from this research. 

\section*{Acknowledgments}

This research is supported by the National Research Foundation, Singapore under its International Research Centres in Singapore Funding Initiative. Any opinions, findings \& conclusions or recommendations expressed in this material are those of the author(s) and do not reflect the views of National Research Foundation, Singapore.

\bibliography{anthology,eacl2021}
\bibliographystyle{acl_natbib}

\end{document}

%% file: pics/SDnoi.tikz
\resizebox{150pt}{123pt}{
\begin{tikzpicture}
    \begin{axis}[
      xmin=1, xmax=15,
      ymin=0, ymax=1,
      ymajorgrids=true,
      grid style=dashed,
      xlabel={\textcolor{white}{Stage}}, 
      ylabel={Ratio of [NOI] tokens}
    ]

    \addplot[color=blue] coordinates {(1,0.52)(2,0.62)(3,0.726)(4,0.817)(5,0.879)(6,0.922)(7,0.949)(8,0.967)(9,0.978)(10,0.985)(11,0.989)(12,0.991)(13,0.992)(14,0.993)(15,0.994)(16,0.993)(17,0.994)(18,0.995)(19,0.995)(20,0.995)(21,0.995)(22,0.995)(23,0.996)(24,0.996)(25,0.995)(26,0.996)(27,0.996)(28,0.997)(29,0.996)(30,0.996)(31,0.996)(32,0.996)(33,0.996)(34,0.997)(35,0.997)(36,0.997)(37,0.997)(38,0.997)(39,0.997)(40,0.997)(41,0.999)(42,0.996)(43,0.996)(44,0.996)(45,0.996)(46,0.996)(47,0.996)(48,0.996)(49,0.996)(50,0.996)(51,0.996)(52,0.996)(53,0.996)(54,0.996)(55,0.996)(56,0.996)(57,0.996)(58,0.996)(59,0.996)(60,0.996)(61,0.996)(62,0.996)(63,0.996)(64,0.996)(65,0.996)(66,1.0)};
    
    \addplot[color=orange] coordinates {(1,0.52)(2,0.565)(3,0.594)(4,0.749)(5,0.906)(6,0.968)(7,0.985)(8,0.992)(9,0.994)(10,0.995)(11,0.995)(12,0.995)(13,0.998)(14,0.998)(15,1.0)};
    
    \addplot[color=green] coordinates {(1,0.937)(2,0.846)(3,0.7)(4,0.528)(5,0.452)(6,0.747)(7,0.947)(8,1.0)};
    
    \addplot[name path=bottom_up_high,color=blue!70] coordinates {(1,0.604)(2,0.726)(3,0.816)(4,0.889)(5,0.936)(6,0.966)(7,0.984)(8,0.994)(9,0.999)(10,1.001)(11,1.002)(12,1.002)(13,1.002)(14,1.002)(15,1.002)(16,1.002)(17,1.003)(18,1.001)(19,1.0)(20,1.002)(21,1.002)(22,1.001)(23,1.001)(24,1.001)(25,0.999)(26,0.999)(27,1.0)(28,0.999)(29,0.997)(30,0.997)(31,0.997)(32,0.997)(33,0.997)(34,0.997)(35,0.997)(36,0.997)(37,0.997)(38,0.997)(39,0.997)(40,0.997)(41,1.001)(42,0.996)(43,0.996)(44,0.996)(45,0.996)(46,0.996)(47,0.996)(48,0.996)(49,0.996)(50,0.996)(51,0.996)(52,0.996)(53,0.996)(54,0.996)(55,0.996)(56,0.996)(57,0.996)(58,0.996)(59,0.996)(60,0.996)(61,0.996)(62,0.996)(63,0.996)(64,0.996)(65,0.996)(66,1.0)};
    \addplot[name path=bottom_up_low,color=blue!70] coordinates {(1,0.437)(2,0.514)(3,0.635)(4,0.745)(5,0.823)(6,0.878)(7,0.915)(8,0.941)(9,0.957)(10,0.968)(11,0.976)(12,0.98)(13,0.982)(14,0.984)(15,0.985)(16,0.985)(17,0.985)(18,0.988)(19,0.99)(20,0.988)(21,0.987)(22,0.99)(23,0.99)(24,0.992)(25,0.991)(26,0.993)(27,0.992)(28,0.995)(29,0.996)(30,0.996)(31,0.996)(32,0.996)(33,0.996)(34,0.996)(35,0.996)(36,0.996)(37,0.996)(38,0.996)(39,0.996)(40,0.996)(41,0.996)(42,0.996)(43,0.996)(44,0.996)(45,0.996)(46,0.996)(47,0.996)(48,0.996)(49,0.996)(50,0.996)(51,0.996)(52,0.996)(53,0.996)(54,0.996)(55,0.996)(56,0.996)(57,0.996)(58,0.996)(59,0.996)(60,0.996)(61,0.996)(62,0.996)(63,0.996)(64,0.996)(65,0.996)(66,1.0)};
    \addplot[blue!50,fill opacity=0.5] fill between[of=bottom_up_high and bottom_up_low];
    
    \addplot[name path=bottom_up_sftmx_high,color=orange!70] coordinates {(1,0.604)(2,0.681)(3,0.723)(4,0.874)(5,0.987)(6,1.01)(7,1.005)(8,1.004)(9,1.004)(10,1.003)(11,1.003)(12,1.003)(13,1.002)(14,1.002)(15,1.0)};
    \addplot[name path=bottom_up_sftmx_low,color=orange!70] coordinates {(1,0.437)(2,0.449)(3,0.466)(4,0.624)(5,0.825)(6,0.926)(7,0.965)(8,0.979)(9,0.983)(10,0.986)(11,0.987)(12,0.988)(13,0.994)(14,0.994)(15,1.0)};
    \addplot[orange!50,fill opacity=0.5] fill between[of=bottom_up_sftmx_high and bottom_up_sftmx_low];

    \addplot[name path=top_down_high,color=green!70] coordinates {(1,0.996)(2,0.941)(3,0.82)(4,0.66)(5,0.699)(6,1.099)(7,1.146)(8,1.0)};
    \addplot[name path=top_down_low,color=green!70] coordinates {(1,0.878)(2,0.75)(3,0.58)(4,0.395)(5,0.205)(6,0.394)(7,0.749)(8,1.0)};
    \addplot[green!50,fill opacity=0.5] fill between[of=top_down_high and top_down_low];

    \end{axis}
\end{tikzpicture}

}

%% file: pics/SMnoi.tikz
\resizebox{150pt}{123pt}{
\begin{tikzpicture}
    \begin{axis}[
    xmin=1, xmax=15,
      ymin=0, ymax=1,
      ymajorgrids=true,
      grid style=dashed,
      xlabel={Stage}, 
      ylabel={\textcolor{white}{$\#$Inserted tokens}}
    ]

    \addplot[color=blue] coordinates {(1,0.463)(2,0.544)(3,0.663)(4,0.77)(5,0.845)(6,0.897)(7,0.933)(8,0.956)(9,0.97)(10,0.978)(11,0.983)(12,0.986)(13,0.988)(14,0.99)(15,0.991)(16,0.991)(17,0.992)(18,0.99)(19,0.992)(20,0.993)(21,0.992)(22,0.992)(23,0.993)(24,0.993)(25,0.993)(26,0.993)(27,0.992)(28,0.994)(29,0.991)(30,0.992)(31,0.994)(32,0.992)(33,0.992)(34,0.994)(35,0.994)(36,0.993)(37,0.993)(38,0.993)(39,0.993)(40,0.999)(41,0.994)(42,0.994)(43,0.994)(44,0.994)(45,0.994)(46,0.994)(47,0.994)(48,0.994)(49,0.994)(50,1.0)};
    \addplot[color=orange] coordinates {(1,0.463)(2,0.478)(3,0.516)(4,0.671)(5,0.861)(6,0.951)(7,0.978)(8,0.987)(9,0.991)(10,0.993)(11,0.994)(12,0.997)(13,0.997)(14,0.996)(15,0.995)(16,1.0)};
    
    \addplot[color=green] coordinates {(1,0.905)(2,0.799)(3,0.63)(4,0.452)(5,0.377)(6,0.687)(7,0.914)(8,0.989)(9,1.0)};
    
    \addplot[name path=bottom_up_high,color=blue!70] coordinates {(1,0.561)(2,0.659)(3,0.761)(4,0.853)(5,0.917)(6,0.955)(7,0.977)(8,0.99)(9,0.997)(10,1.0)(11,1.002)(12,1.002)(13,1.002)(14,1.002)(15,1.002)(16,1.002)(17,1.003)(18,1.001)(19,1.0)(20,1.002)(21,1.001)(22,1.0)(23,1.0)(24,0.999)(25,0.997)(26,0.996)(27,0.996)(28,1.003)(29,1.001)(30,0.997)(31,1.0)(32,0.994)(33,0.994)(34,0.996)(35,0.997)(36,0.994)(37,0.994)(38,0.994)(39,0.994)(40,1.002)(41,0.994)(42,0.994)(43,0.994)(44,0.994)(45,0.994)(46,0.994)(47,0.994)(48,0.994)(49,0.994)(50,1.0)};
    \addplot[name path=bottom_up_low,color=blue!70] coordinates {(1,0.366)(2,0.43)(3,0.565)(4,0.686)(5,0.774)(6,0.84)(7,0.889)(8,0.922)(9,0.943)(10,0.957)(11,0.965)(12,0.97)(13,0.974)(14,0.978)(15,0.979)(16,0.979)(17,0.98)(18,0.979)(19,0.984)(20,0.983)(21,0.984)(22,0.985)(23,0.986)(24,0.987)(25,0.989)(26,0.99)(27,0.989)(28,0.985)(29,0.981)(30,0.987)(31,0.988)(32,0.991)(33,0.991)(34,0.991)(35,0.991)(36,0.993)(37,0.993)(38,0.993)(39,0.993)(40,0.996)(41,0.994)(42,0.994)(43,0.994)(44,0.994)(45,0.994)(46,0.994)(47,0.994)(48,0.994)(49,0.994)(50,1.0)};
    \addplot[blue!50,fill opacity=0.5] fill between[of=bottom_up_high and bottom_up_low];
    
    \addplot[name path=bottom_up_sftmx_high,color=orange!70] coordinates {(1,0.561)(2,0.599)(3,0.652)(4,0.816)(5,0.968)(6,1.01)(7,1.005)(8,1.005)(9,1.004)(10,1.003)(11,1.003)(12,1.003)(13,1.003)(14,1.004)(15,1.0)(16,1.0)};
    \addplot[name path=bottom_up_sftmx_low,color=orange!70] coordinates {(1,0.366)(2,0.357)(3,0.38)(4,0.526)(5,0.754)(6,0.892)(7,0.951)(8,0.97)(9,0.978)(10,0.982)(11,0.986)(12,0.99)(13,0.99)(14,0.989)(15,0.99)(16,1.0)};
    \addplot[orange!50,fill opacity=0.5] fill between[of=bottom_up_sftmx_high and bottom_up_sftmx_low];
    
    \addplot[name path=top_down_high,color=green!70] coordinates {(1,0.977)(2,0.909)(3,0.755)(4,0.577)(5,0.616)(6,1.076)(7,1.17)(8,1.086)(9,1.0)};
    \addplot[name path=top_down_low,color=green!70] coordinates {(1,0.833)(2,0.689)(3,0.505)(4,0.328)(5,0.139)(6,0.299)(7,0.659)(8,0.892)(9,1.0)};
    \addplot[green!50,fill opacity=0.5] fill between[of=top_down_high and top_down_low];

    \end{axis}
\end{tikzpicture}
}

%% file: pics/CoNLLnoi.tikz
\resizebox{150pt}{123pt}{
\begin{tikzpicture}
    \begin{axis}[
      xmin=1, xmax=15,
      ymin=0, ymax=1,
      ymajorgrids=true,
      grid style=dashed,
      legend pos=south east,
      xlabel={\textcolor{white}{Stage}}, 
      ylabel={\textcolor{white}{$\#$Inserted tokens}}
    ]

    \addplot[color=blue] coordinates {(1,0.632)(2,0.695)(3,0.763)(4,0.822)(5,0.864)(6,0.901)(7,0.927)(8,0.945)(9,0.957)(10,0.966)(11,0.972)(12,0.976)(13,0.98)(14,0.983)(15,0.985)(16,0.987)(17,0.988)(18,0.989)(19,0.989)(20,0.991)(21,0.991)(22,0.991)(23,0.991)(24,0.992)(25,0.992)(26,0.992)(27,0.993)(28,0.994)(29,0.993)(30,0.991)(31,0.991)(32,0.99)(33,0.99)(34,0.99)(35,0.991)(36,0.993)(37,0.992)(38,0.993)(39,0.994)(40,0.995)(41,0.995)(42,0.996)(43,0.996)(44,0.995)(45,0.994)(46,0.995)(47,0.995)(48,0.996)(49,0.996)(50,0.997)(51,0.996)(52,0.997)(53,0.996)(54,0.997)(55,0.997)(56,0.997)(57,0.997)(58,0.997)(59,0.998)(60,0.997)(61,0.997)(62,0.997)(63,0.997)(64,0.998)(65,0.997)(66,0.997)(67,0.997)(68,0.998)(69,0.997)(70,0.997)(71,0.998)(72,0.996)(73,0.996)(74,0.996)(75,0.996)(76,0.996)(77,0.996)(78,0.996)(79,0.997)(80,0.997)(81,0.997)(82,0.997)(83,0.997)(84,0.997)(85,0.997)(86,0.997)(87,0.997)(88,0.997)(89,0.997)(90,0.997)(91,0.997)(92,0.997)(93,0.997)(94,0.997)(95,0.997)(96,0.997)(97,0.997)(98,0.997)(99,1.0)};
    \addlegendentry{\textsc{Greedy Enconter}}
    
    \addplot[color=orange] coordinates {(1,0.632)(2,0.597)(3,0.585)(4,0.73)(5,0.859)(6,0.919)(7,0.955)(8,0.974)(9,0.982)(10,0.986)(11,0.989)(12,0.991)(13,0.991)(14,0.991)(15,0.992)(16,0.994)(17,0.996)(18,0.996)(19,0.994)(20,0.996)(21,0.997)(22,0.994)(23,1.0)};
    \addlegendentry{\textsc{BBT-Enconter}}
    
    \addplot[color=green] coordinates {(1,0.95)(2,0.881)(3,0.753)(4,0.583)(5,0.512)(6,0.586)(7,0.754)(8,0.837)(9,0.97)(10,1.0)};
    \addlegendentry{\textsc{Pointer-E}}
    
    \addplot[name path=bottom_up_high,color=blue!70] coordinates {(1,0.728)(2,0.797)(3,0.847)(4,0.891)(5,0.921)(6,0.954)(7,0.975)(8,0.987)(9,0.992)(10,0.995)(11,0.998)(12,0.999)(13,1.0)(14,1.0)(15,1.001)(16,1.001)(17,1.001)(18,1.002)(19,1.002)(20,1.002)(21,1.001)(22,1.0)(23,1.0)(24,0.999)(25,0.999)(26,0.999)(27,0.999)(28,0.999)(29,1.0)(30,1.0)(31,1.001)(32,1.001)(33,1.002)(34,1.0)(35,0.999)(36,0.999)(37,0.999)(38,0.998)(39,0.998)(40,0.999)(41,0.999)(42,1.0)(43,1.001)(44,0.999)(45,0.999)(46,0.999)(47,0.998)(48,1.0)(49,0.999)(50,0.999)(51,0.999)(52,0.999)(53,0.998)(54,0.998)(55,0.998)(56,0.998)(57,0.998)(58,0.998)(59,0.999)(60,0.998)(61,0.998)(62,0.998)(63,0.998)(64,0.999)(65,0.998)(66,0.998)(67,0.998)(68,1.0)(69,0.997)(70,0.997)(71,0.998)(72,0.996)(73,0.996)(74,0.996)(75,0.996)(76,0.996)(77,0.996)(78,0.996)(79,0.997)(80,0.997)(81,0.997)(82,0.997)(83,0.997)(84,0.997)(85,0.997)(86,0.997)(87,0.997)(88,0.997)(89,0.997)(90,0.997)(91,0.997)(92,0.997)(93,0.997)(94,0.997)(95,0.997)(96,0.997)(97,0.997)(98,0.997)(99,1.0)};
    \addplot[name path=bottom_up_low,color=blue!70] coordinates {(1,0.536)(2,0.594)(3,0.678)(4,0.752)(5,0.806)(6,0.849)(7,0.879)(8,0.903)(9,0.923)(10,0.937)(11,0.947)(12,0.954)(13,0.96)(14,0.966)(15,0.97)(16,0.972)(17,0.975)(18,0.976)(19,0.977)(20,0.98)(21,0.981)(22,0.981)(23,0.983)(24,0.984)(25,0.986)(26,0.986)(27,0.987)(28,0.988)(29,0.985)(30,0.982)(31,0.982)(32,0.98)(33,0.978)(34,0.98)(35,0.982)(36,0.986)(37,0.986)(38,0.989)(39,0.989)(40,0.99)(41,0.991)(42,0.992)(43,0.991)(44,0.991)(45,0.989)(46,0.991)(47,0.991)(48,0.992)(49,0.994)(50,0.994)(51,0.993)(52,0.995)(53,0.995)(54,0.996)(55,0.996)(56,0.996)(57,0.997)(58,0.997)(59,0.996)(60,0.996)(61,0.997)(62,0.997)(63,0.997)(64,0.996)(65,0.996)(66,0.996)(67,0.996)(68,0.996)(69,0.997)(70,0.997)(71,0.998)(72,0.996)(73,0.996)(74,0.996)(75,0.996)(76,0.996)(77,0.996)(78,0.996)(79,0.997)(80,0.997)(81,0.997)(82,0.997)(83,0.997)(84,0.997)(85,0.997)(86,0.997)(87,0.997)(88,0.997)(89,0.997)(90,0.997)(91,0.997)(92,0.997)(93,0.997)(94,0.997)(95,0.997)(96,0.997)(97,0.997)(98,0.997)(99,1.0)};
    \addplot[blue!50,fill opacity=0.5] fill between[of=bottom_up_high and bottom_up_low];
    
    \addplot[name path=bottom_up_sftmx_high,color=orange!70] coordinates {(1,0.728)(2,0.722)(3,0.723)(4,0.873)(5,0.98)(6,1.018)(7,1.02)(8,1.017)(9,1.006)(10,1.003)(11,1.001)(12,1.0)(13,1.0)(14,0.999)(15,0.999)(16,0.999)(17,1.001)(18,1.001)(19,0.999)(20,1.001)(21,0.997)(22,0.994)(23,1.0)};
    \addplot[name path=bottom_up_sftmx_low,color=orange!70] coordinates {(1,0.536)(2,0.472)(3,0.448)(4,0.587)(5,0.739)(6,0.82)(7,0.889)(8,0.93)(9,0.957)(10,0.969)(11,0.977)(12,0.981)(13,0.982)(14,0.983)(15,0.985)(16,0.989)(17,0.992)(18,0.991)(19,0.988)(20,0.991)(21,0.997)(22,0.994)(23,1.0)};
    \addplot[orange!50,fill opacity=0.5] fill between[of=bottom_up_sftmx_high and bottom_up_sftmx_low];

    \addplot[name path=top_down_high,color=green!70] coordinates {(1,0.991)(2,0.956)(3,0.861)(4,0.715)(5,0.746)(6,0.939)(7,1.119)(8,1.175)(9,1.134)(10,1.0)};
    \addplot[name path=top_down_low,color=green!70] coordinates {(1,0.908)(2,0.806)(3,0.645)(4,0.452)(5,0.278)(6,0.233)(7,0.389)(8,0.499)(9,0.806)(10,1.0)};
    \addplot[green!50,fill opacity=0.5] fill between[of=top_down_high and top_down_low];

    \end{axis}
\end{tikzpicture}
}

%% file: eacl2021.bbl
\begin{thebibliography}{32}
\expandafter\ifx\csname natexlab\endcsname\relax\def\natexlab#1{#1}\fi

\bibitem[{Bahdanau et~al.(2015)Bahdanau, Cho, and Bengio}]{bahdanau2015neural}
Dzmitry Bahdanau, Kyunghyun Cho, and Yoshua Bengio. 2015.
\newblock Neural machine translation by jointly learning to align and
  translate.
\newblock In \emph{3rd International Conference on Learning Representations,
  ICLR 2015}.

\bibitem[{Campos et~al.(2020)Campos, Mangaravite, Pasquali, Jorge, Nunes, and
  Jatowt}]{campos2020yake}
Ricardo Campos, V{\'\i}tor Mangaravite, Arian Pasquali, Al{\'\i}pio Jorge,
  C{\'e}lia Nunes, and Adam Jatowt. 2020.
\newblock Yake! keyword extraction from single documents using multiple local
  features.
\newblock \emph{Information Sciences}, 509:257--289.

\bibitem[{Chan et~al.(2019)Chan, Kitaev, Guu, Stern, and
  Uszkoreit}]{chan2019kermit}
William Chan, Nikita Kitaev, Kelvin Guu, Mitchell Stern, and Jakob Uszkoreit.
  2019.
\newblock Kermit: Generative insertion-based modeling for sequences.
\newblock \emph{arXiv preprint arXiv:1906.01604}.

\bibitem[{Chopra et~al.(2016)Chopra, Auli, and Rush}]{chopra2016abstractive}
Sumit Chopra, Michael Auli, and Alexander~M Rush. 2016.
\newblock Abstractive sentence summarization with attentive recurrent neural
  networks.
\newblock In \emph{Proceedings of the 2016 Conference of the North American
  Chapter of the Association for Computational Linguistics: Human Language
  Technologies}, pages 93--98.

\bibitem[{Dai and Le(2015)}]{dai2015semi}
Andrew~M Dai and Quoc~V Le. 2015.
\newblock Semi-supervised sequence learning.
\newblock In \emph{Advances in neural information processing systems}, pages
  3079--3087.

\bibitem[{Dathathri et~al.(2020)Dathathri, Madotto, Lan, Hung, Frank, Molino,
  Yosinski, and Liu}]{Dathathri2020Plug}
Sumanth Dathathri, Andrea Madotto, Janice Lan, Jane Hung, Eric Frank, Piero
  Molino, Jason Yosinski, and Rosanne Liu. 2020.
\newblock \href {https://openreview.net/forum?id=H1edEyBKDS} {Plug and play
  language models: A simple approach to controlled text generation}.
\newblock In \emph{International Conference on Learning Representations}.

\bibitem[{Devlin et~al.(2019)Devlin, Chang, Lee, and
  Toutanova}]{devlin-etal-2019-bert}
Jacob Devlin, Ming-Wei Chang, Kenton Lee, and Kristina Toutanova. 2019.
\newblock \href {https://doi.org/10.18653/v1/N19-1423} {{BERT}: Pre-training of
  deep bidirectional transformers for language understanding}.
\newblock In \emph{Proceedings of the 2019 Conference of the North {A}merican
  Chapter of the Association for Computational Linguistics: Human Language
  Technologies, Volume 1 (Long and Short Papers)}, pages 4171--4186,
  Minneapolis, Minnesota. Association for Computational Linguistics.

\bibitem[{Doddington(2002)}]{doddington2002automatic}
George Doddington. 2002.
\newblock Automatic evaluation of machine translation quality using n-gram
  co-occurrence statistics.
\newblock In \emph{Proceedings of the second international conference on Human
  Language Technology Research}, pages 138--145.

\bibitem[{Gatt and Krahmer(2018)}]{gatt2018survey}
Albert Gatt and Emiel Krahmer. 2018.
\newblock Survey of the state of the art in natural language generation: Core
  tasks, applications and evaluation.
\newblock \emph{Journal of Artificial Intelligence Research}, 61:65--170.

\bibitem[{Gries(1982)}]{gries1982note}
David Gries. 1982.
\newblock A note on a standard strategy for developing loop invariants and
  loops.
\newblock \emph{Science of Computer Programming}, 2(3):207--214.

\bibitem[{Gu et~al.(2019)Gu, Wang, and Zhao}]{gu2019levenshtein}
Jiatao Gu, Changhan Wang, and Junbo Zhao. 2019.
\newblock Levenshtein transformer.
\newblock In \emph{Advances in Neural Information Processing Systems}, pages
  11181--11191.

\bibitem[{H.~Lee et~al.(2020)H.~Lee, Shu, Achananuparp, Prasetyo, Liu, Lim, and
  Varshney}]{h2020recipegpt}
Helena H.~Lee, Ke~Shu, Palakorn Achananuparp, Philips~Kokoh Prasetyo, Yue Liu,
  Ee-Peng Lim, and Lav~R Varshney. 2020.
\newblock Recipegpt: Generative pre-training based cooking recipe generation
  and evaluation system.
\newblock In \emph{Companion Proceedings of the Web Conference 2020}, pages
  181--184.

\bibitem[{Hokamp and Liu(2017)}]{hokamp2017lexically}
Chris Hokamp and Qun Liu. 2017.
\newblock Lexically constrained decoding for sequence generation using grid
  beam search.
\newblock In \emph{Proceedings of the 55th Annual Meeting of the Association
  for Computational Linguistics (Volume 1: Long Papers)}, pages 1535--1546.

\bibitem[{Hu et~al.(2019)Hu, Khayrallah, Culkin, Xia, Chen, Post, and
  Van~Durme}]{hu2019improved}
J~Edward Hu, Huda Khayrallah, Ryan Culkin, Patrick Xia, Tongfei Chen, Matt
  Post, and Benjamin Van~Durme. 2019.
\newblock Improved lexically constrained decoding for translation and
  monolingual rewriting.
\newblock In \emph{Proceedings of the 2019 Conference of the North American
  Chapter of the Association for Computational Linguistics: Human Language
  Technologies, Volume 1 (Long and Short Papers)}, pages 839--850.

\bibitem[{Keskar et~al.(2019)Keskar, McCann, Varshney, Xiong, and
  Socher}]{keskarCTRL2019}
Nitish~Shirish Keskar, Bryan McCann, Lav Varshney, Caiming Xiong, and Richard
  Socher. 2019.
\newblock {CTRL - A Conditional Transformer Language Model for Controllable
  Generation}.
\newblock \emph{arXiv preprint arXiv:1909.05858}.

\bibitem[{Lavie and Agarwal(2007)}]{lavie2007meteor}
Alon Lavie and Abhaya Agarwal. 2007.
\newblock Meteor: An automatic metric for mt evaluation with high levels of
  correlation with human judgments.
\newblock In \emph{Proceedings of the second workshop on statistical machine
  translation}, pages 228--231.

\bibitem[{Li et~al.(2016)Li, Galley, Brockett, Gao, and
  Dolan}]{li2016diversity}
Jiwei Li, Michel Galley, Chris Brockett, Jianfeng Gao, and Bill Dolan. 2016.
\newblock A diversity-promoting objective function for neural conversation
  models.
\newblock In \emph{Proceedings of the 2016 Conference of the North American
  Chapter of the Association for Computational Linguistics: Human Language
  Technologies}, pages 110--119.

\bibitem[{Miao et~al.(2019)Miao, Zhou, Mou, Yan, and Li}]{miao2019cgmh}
Ning Miao, Hao Zhou, Lili Mou, Rui Yan, and Lei Li. 2019.
\newblock Cgmh: Constrained sentence generation by metropolis-hastings
  sampling.
\newblock In \emph{Proceedings of the AAAI Conference on Artificial
  Intelligence}, volume~33, pages 6834--6842.

\bibitem[{Papineni et~al.(2002)Papineni, Roukos, Ward, and
  Zhu}]{papineni2002bleu}
Kishore Papineni, Salim Roukos, Todd Ward, and Wei-Jing Zhu. 2002.
\newblock Bleu: a method for automatic evaluation of machine translation.
\newblock In \emph{Proceedings of the 40th annual meeting of the Association
  for Computational Linguistics}, pages 311--318.

\bibitem[{Peters et~al.(2018)Peters, Neumann, Iyyer, Gardner, Clark, Lee, and
  Zettlemoyer}]{peters2018deep}
Matthew Peters, Mark Neumann, Mohit Iyyer, Matt Gardner, Christopher Clark,
  Kenton Lee, and Luke Zettlemoyer. 2018.
\newblock Deep contextualized word representations.
\newblock In \emph{Proceedings of the 2018 Conference of the North American
  Chapter of the Association for Computational Linguistics: Human Language
  Technologies, Volume 1 (Long Papers)}, pages 2227--2237.

\bibitem[{Post and Vilar(2018)}]{post2018fast}
Matt Post and David Vilar. 2018.
\newblock Fast lexically constrained decoding with dynamic beam allocation for
  neural machine translation.
\newblock In \emph{Proceedings of the 2018 Conference of the North American
  Chapter of the Association for Computational Linguistics: Human Language
  Technologies, Volume 1 (Long Papers)}, pages 1314--1324.

\bibitem[{Qin et~al.(2019)Qin, Galley, Brockett, Liu, Gao, Dolan, Choi, and
  Gao}]{qin2019conversing}
Lianhui Qin, Michel Galley, Chris Brockett, Xiaodong Liu, Xiang Gao, Bill
  Dolan, Yejin Choi, and Jianfeng Gao. 2019.
\newblock Conversing by reading: Contentful neural conversation with on-demand
  machine reading.
\newblock In \emph{Proceedings of the 57th Annual Meeting of the Association
  for Computational Linguistics}, pages 5427--5436.

\bibitem[{Radford(2018)}]{Radford2018ImprovingLU}
A.~Radford. 2018.
\newblock Improving language understanding by generative pre-training.

\bibitem[{Radford et~al.(2019)Radford, Wu, Child, Luan, Amodei, and
  Sutskever}]{radford2019language}
Alec Radford, Jeff Wu, Rewon Child, David Luan, Dario Amodei, and Ilya
  Sutskever. 2019.
\newblock Language models are unsupervised multitask learners.

\bibitem[{Stern et~al.(2019)Stern, Chan, Kiros, and
  Uszkoreit}]{stern2019insertion}
Mitchell Stern, William Chan, Jamie Kiros, and Jakob Uszkoreit. 2019.
\newblock Insertion transformer: Flexible sequence generation via insertion
  operations.
\newblock In \emph{International Conference on Machine Learning}, pages
  5976--5985.

\bibitem[{Tang et~al.(2019)Tang, Zhao, Xiong, Liang, Xing, and
  Hu}]{tang2019target}
Jianheng Tang, Tiancheng Zhao, Chenyan Xiong, Xiaodan Liang, Eric Xing, and
  Zhiting Hu. 2019.
\newblock Target-guided open-domain conversation.
\newblock In \emph{Proceedings of the 57th Annual Meeting of the Association
  for Computational Linguistics}, pages 5624--5634.

\bibitem[{Tjong Kim~Sang and De~Meulder(2003)}]{tjong2003introduction}
Erik~F. Tjong Kim~Sang and Fien De~Meulder. 2003.
\newblock \href {https://www.aclweb.org/anthology/W03-0419} {Introduction to
  the {C}o{NLL}-2003 shared task: Language-independent named entity
  recognition}.
\newblock In \emph{Proceedings of the Seventh Conference on Natural Language
  Learning at {HLT}-{NAACL} 2003}, pages 142--147.

\bibitem[{Welleck et~al.(2019)Welleck, Brantley, Daum{\'e}, and
  Cho}]{welleck2019non}
Sean Welleck, Kiant{\'e} Brantley, Hal Daum{\'e}, and Kyunghyun Cho. 2019.
\newblock Non-monotonic sequential text generation.
\newblock In \emph{36th International Conference on Machine Learning, ICML
  2019}, pages 11656--11676. International Machine Learning Society (IMLS).

\bibitem[{Wu et~al.(2016)Wu, Schuster, Chen, Le, Norouzi, Macherey, Krikun,
  Cao, Gao, Macherey et~al.}]{wu2016google}
Yonghui Wu, Mike Schuster, Zhifeng Chen, Quoc~V Le, Mohammad Norouzi, Wolfgang
  Macherey, Maxim Krikun, Yuan Cao, Qin Gao, Klaus Macherey, et~al. 2016.
\newblock Google's neural machine translation system: Bridging the gap between
  human and machine translation.
\newblock \emph{arXiv preprint arXiv:1609.08144}.

\bibitem[{Zhang et~al.(2018)Zhang, Galley, Gao, Gan, Li, Brockett, and
  Dolan}]{zhang2018generating}
Yizhe Zhang, Michel Galley, Jianfeng Gao, Zhe Gan, Xiujun Li, Chris Brockett,
  and Bill Dolan. 2018.
\newblock Generating informative and diverse conversational responses via
  adversarial information maximization.
\newblock In \emph{Advances in Neural Information Processing Systems}, pages
  1810--1820.

\bibitem[{Zhang et~al.(2020)Zhang, Wang, Li, Gan, Brockett, and
  Dolan}]{zhang2020pointer}
Yizhe Zhang, Guoyin Wang, Chunyuan Li, Zhe Gan, Chris Brockett, and Bill Dolan.
  2020.
\newblock Pointer: Constrained text generation via insertion-based generative
  pre-training.
\newblock \emph{arXiv preprint arXiv:2005.00558}.

\bibitem[{Zugarini et~al.(2019)Zugarini, Melacci, and
  Maggini}]{zugarini2019neural}
Andrea Zugarini, Stefano Melacci, and Marco Maggini. 2019.
\newblock Neural poetry: Learning to generate poems using syllables.
\newblock In \emph{International Conference on Artificial Neural Networks},
  pages 313--325. Springer.

\end{thebibliography}
